# ACCURATE LOCALIZATION IN DENSE URBAN AREA USING GOOGLE STREET VIEW IMAGES

*Mahdi Salarian      Andrea Manavella    Rashid Ansari*
*Email: msalar2@uic.edu    Email  mana1990@hotmail.it   Email: ransa@uic.edu*
*School of Electrical and Computer Engineering*
*University of Illinois at Chicago*
*Chicago, IL*

## ABSTRACT

Accurate information about the location and orientation of a camera in mobile devices is central to the utilization of location-based services (LBS). Most of such mobile devices rely on GPS data but this data is subject to inaccuracy due to imperfections in the quality of the signal provided by satellites. This shortcoming has spurred the research into improving the accuracy of localization. Since mobile devices have camera, a major thrust of this research has been seeks to acquire the local scene and apply image retrieval techniques by querying a GPS-tagged image database to find the best match for the acquired scene.. The techniques are however computationally demanding and unsuitable for real-time applications such as assistive technology for navigation by the blind and visually impaired which motivated out work. To overcome the high complexity of those techniques, we investigated the use of inertial sensors as an aid in image-retrieval-based approach. Armed with information of media other than images, such as data from the GPS module along with orientation sensors such as accelerometer and gyro, we sought to limit the size of the image set to c search for the best match. Specifically, data from the orientation sensors along with Dilution of precision (DOP) from GPS are used to find the angle of view and estimation of position. We present analysis of the reduction in the image set size for the search as well as simulations to demonstrate the effectiveness in a fast implementation with 98%  Estimated Position Error.

*Index Terms*— mobile device, LBS, Geotagged-images, TF-IDF, GSW.

## 1. INTRODUCTION

Finding accurate location of a device is a really interesting area of research. Traditional approaches are using GPS module or distance from cellular towers to find location. Although GPS are usually accurate enough but it's performance depends on satellite navigation system to get sufficient precision. In the other hand, GPS is usually accurate when it has a clear view of sky to get signal at least from 4 satellites.  This accuracy is really crucial for people who are more dependent to LBS technologies like blind and visual impaired people. Although in most of the time GPS are used in vehicle and show great performance, but for pedestrian who are moving by velocity less than 8 kilometer in sidewalk of street where there are tall wall and building, finding accurate localization is really difficult. This difficulties are going to be worst in dens region of cities. For example reports claim that the average localization errors of mobile phone GPS are in the range of 50-100 meter [1] so some of researchers focused on using inertial sensor or camera to address GPS unreliability. Having camera, they tried to figure out whether is it possible to get accurate positions of a query image generated by the camera using computer vision techniques. In fact those approaches are based on searching the best match for a query image in a database of Geo-reference images with accurate GPS coordination. One of these references is Google street view images. Image of GSW are acquired by precise capturing and mapping device and can be used as a reference image in database [2]. By having the best match it is feasible to estimate coordination of query that is taken by our hand held or wearable devices. The key idea of this approaches is scene recognition using feature extraction and matching techniques. There are numbers of existing successful approaches for whole process that are among image retrieval category. To do this scale invariant feature such as SIFT [3], SURF [4] or Dense SIFT are used. This feature extraction schemes show better performance than prior algorithms such as MSER [9]. Having huge number of images in database results huge number of features that needs to be searched so many times for each query. So prior knowledge of position is used in some researches to narrow search space. For example in [5] a city scale dataset is partitioned with special algorithm to result in couple of smaller partitions. As a result system just need to search in smaller partition that is close to estimated location of user. In number of researches people made their own dataset and made it possible for other researchers to use it and compare results. They collect image by their device or just by downloading images from websites like Flicker or even services like Google Street View. In this research we download GSW image directly and use it as a database. Because our goal is finding more accurate position to help pedestrian and especially visual impaired people we just got image that may be seen by a person who is walking in side of street. In next sections we talked about related work. Then our method for making dataset is discussed. Then the

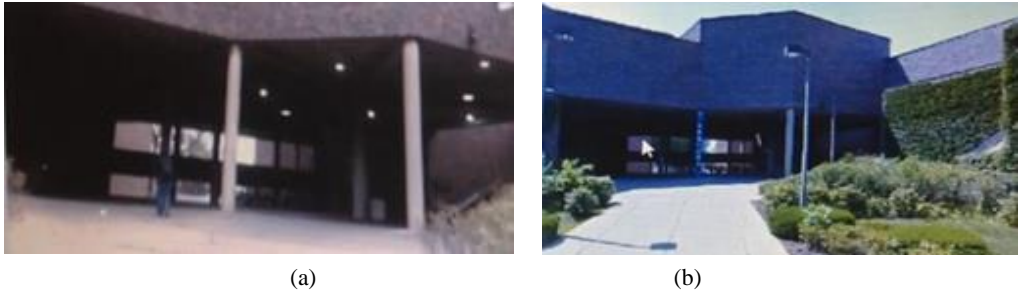

Figure 1. corresponding images from camera and Google street view when GPS can see clear sky
a: image by camera  b:image from Google street view proposed method for confining search plus homology verification step have been evaluated.
Finally some results for real difficult sample where GPS and cell Id approach aren't successful have been shown.

## 3. RELATED WORK

Recent computer vision achievements made it possible to could recognize natural features in outdoor environment. A main application for this capability is using massive number of Geo-tagged images on the internet to find location for a candidate image [6]. There are variety of methods to do this. For instance Retiary and Drummand [7] utilized an edge based method to get street facades based on tree dimensional method. Other researches tried to modify it by using an initializing step by using an accurate GPS antenna [8]. The most efficient and accurate approach is based on features such as scale invariant feature ( SIFT) and its variation. Using this features make it possible for algorithm to reach to reasonable results even when the query is suffering from different illumination or viewpoint issues. Beside these popular and reliable feature extraction methods, MSRE [9] and compressed histogram of gradient have been used in some researches [10]. After feature extraction step for images in database, features of query should be searched among all features of database using CBIR approaches to recognize the best match reference image in the database. One of the best approach that is widely used is Bag Of Features (BOF) approach that in fact barrowed from text retrieval techniques. This method proposed by Zisserman and Sivic [11]. In this method all features descriptions are quantized to visual words with a clustering algorithm like k-means. So an image can be represented by histogram of number of visual words. Each image in database has its own histogram so for finding the best match histogram of query should be compared with all histograms in database to infer which member of database is more similar to query. There are different methods for finding similarity such as just inner product of two BOW vectors. Although similarity measures can be calculated easily by even L1 distance, but inverted file [12] method is used more and showed great performance. Some other researches focused on clustering step to find an efficient quantization technique to assign each feature descriptor to a visual word . For example Hierarchical k-means and Approximate k-means that should be used when we are dealing with huge amount of data. In fact regular k-means can't be used for huge number of images. Moreover, some modifications like soft assignment instead of hard assignment was proposed to compensate incorrectly assignment for a sample feature vector. For instance Philbin [13] used soft assignment for each descriptor to some close cluster centers considering Gaussian function $e^{\frac{-d}{2\delta}}$ where $\delta$ depends on method and application and d is Euclidean distance. From others experience this isn't usually the last step. Actually most of researches that are based on image retrieval techniques select a couple of best matched candidates. The number of candidate for each application is different and vary from 10 to 100. Actually for finding best match there is another step called Homology verification. This step usually utilize popular algorithm of RANSAC [14] to find the best geometric match between query and candidates. In fact this step is going to compensate the weakness of image retrieval schemes based on BOF due to neglecting geometric information of images. All of mention approach are not aware of approximate GPS coordination and camera's orientation. So research like [15] proposed a method to use inertial sensor and BOF to get more accurate result. Since most of mobile device like smartphones and even cameras are equipped by inertial GPS and even Inertial measurement unit (IMU), we thought searching among all reference image in database is not necessary when inaccurate position and orientation of query are exist. In fact in this research we are going to confine the number of image in dataset by considering prior knowledge from location and using Dilution of precision (DOP) along with direction of camera while it is generating query and go directly to homology verification steps to find best match. Although we have different number of candidates and often more than 20 images in next step, it seems it is a practical approach and can be used in real condition. The organization of paper is as a follows: after describing method for generating our dataset in section 2, we go to section 3 that is about our proposed method of local searching and Homology verification step. Experimental results and conclusion are presented in next 2 sections.

## 4. MAKING DATASET

Since the introduction of Google street view, some researchers used its images to make a dataset. This service

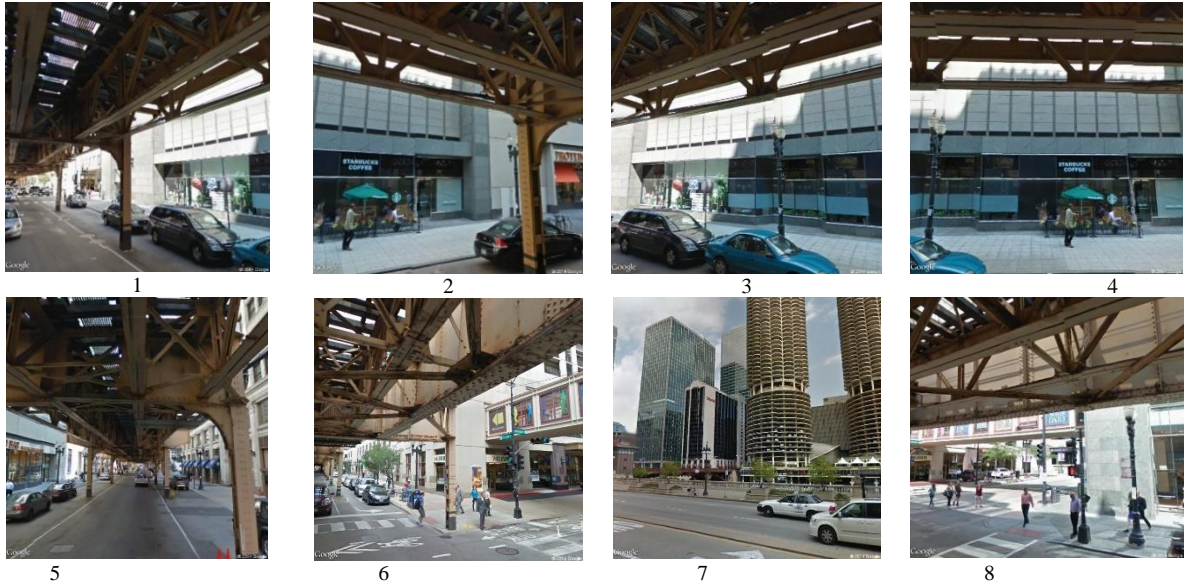
Figure 2. Some candidates for a sample query.

covers most of US cities and more than 10 countries in 4 continents. Although it is free, but in case of making a dataset including thousands of images a request should be sent to Google to get permission for downloading images. Having this permit and utilizing Google street view API make it easy to download image directly from a URL. Despite most of researches that downloaded Panorama image and split it in more than10 parts, our method downloads image for each coordination that Google Street Panorama is exist. So by using a JavaScript we tried to find all coordination in a region specified by a center and radius that have a unique Panorama. Then a URL request such as (1) including latitude, longitude, pitch, yaw, heading and FOV has been sent to get each image. Since the aim of this research is localization, pitch and FOV are assigned to 10 and zero to get images that are similar to view point of a person who is walking. Also for each coordination we considered 12 images that cover every 30 degree by changing heading. Dataset contains Coordination and orientation of all images. This coordination should be accurate because Google vehicles are usually equipped by accurate measurement units. We tried to find by having accurate coordination and orientation parameter such as pitch and yaw whether corresponding image from camera and Google street view are similar or not. Figure 1 shows a sample result when we have almost accurate coordination. As it is clear image from Google Street is from the same scene with a little change. For this research more than 24000 images that is corresponded to more than 2000 position of panorama images are downloaded. For each image sift descriptor is calculated by using Vlfeat library [16].

## 5. PROPOSED METHOD

**5.1.** Our approach consist of two steps. At the first step, a query image should be acquired along with the angle and position. We did it by both cell phone and a hardware designed using GPS module, Camera and IMU. So for each frame of image we are able to save coordination and orientation. Our experiment showed orientation got from sensor is usually enough accurate while coordination may not be accurate enough. It is possible to have an approximation of this inaccuracy using Dilution of precision (DOP) taken from GPS module and calculate Estimated Position Error (EPE).

```
http://maps.googleapis.com/maps/api/streetview?
size=1400x1200&location=',lat,long,
&fov=60&heading=10&pitch=10&sensor=true&key=AIz
aSyAXLYIw71e1c-4e4F9pgTT93OxhI_Qa_WQ
```

Sample URL request for downloading image from Google street view

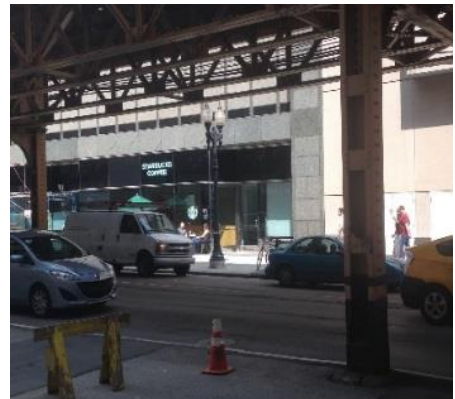
Figure 3. Sample query

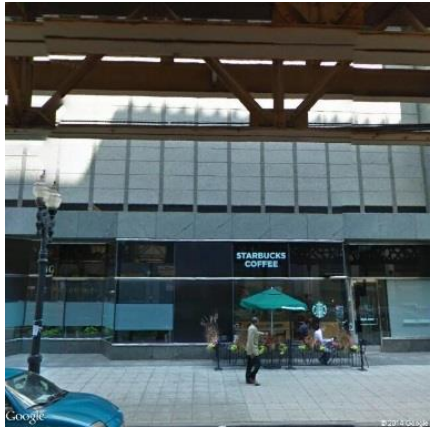 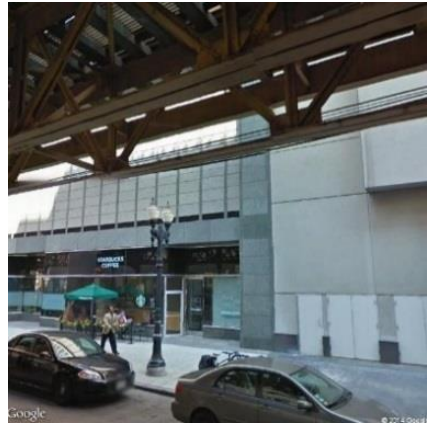

Figure 4    (a) Result for $EPE$ =40 and $K$ =60      (b) Result for $EPE$ =40 and $K$ = 15

This value is actually the radius of a light blue circle around estimated coordination on Google map such as image in figure 4.b. So we decided to consider EPE as parameter to narrow search area. To find EPE we have:

$$EPE(98\% \text{ accuracy}) = 2 \times HDOP \times UERE \quad (3)$$

Where HDOP can be directly extracted from GPS module and $UERE$ is User Equivalent Range Error and is computed in the standard error model tables for GPS. There is this table in [17]. Current GPS use filtering to obtain better accuracy then the value used is 10.2 m. In fact by knowing how the accuracy of result is, we are able to infer how many images should be searched. So in place where sky can be seen easily we don't need to use localization based on image search since data from GPS has sufficient accuracy and reliable for navigation. For finding all image in database located in a neighbor of a coordination (2) is used that calculates distance between two coordination.

$$Distance(I1, I2) = \cos^{-1}(\sin(lat1)\sin(lat2) + \cos(lat1)\cos(lat2)\cos(long2 - long1)) \times R \quad (2)$$

Here R is earth radius that is almost 6371 kilometer. The number of resulting coordinations are directly associated to EPE that representing GPS accuracy. Another parameter is used here is query's orientation. The only images with orientation in the range of $\pm k$ degree have been selected for next step. Proposed algorithm is shown below. Also resulted candidate for example for a query image of figure 3 for 8 closest images in terms of distance are shown in figure 2. In next step all of those candidate images are fed to Homology verification algorithm. Before applying RANSAC, the best match features have been found with considering this criteria:

$$\frac{d(F_i, G_1)}{d(F_i, G_2)} < T$$

Where $d(F_i, G_1)$ is distance between feature $i$ of image $F$ and Closest feature in image $G$, and $d(F_i, G_2)$ is second closest feature in image $G$. Here $T$ is set to 1.8.

## 6. EXPRIMENTAL RESULT

5.2. The procedure for finding best candidates:
Initialization:
**Result**: How to acquire the best candidates
Get GPS and sensors data for generated query;
Calculate the $EPE$ from GPS data
**for** all the images $i$ of $S = \{s_1 s_2, \dots, s_d\}$
$d$=number of images in data set
find $s_n \quad n = 1,..l$ if $Distance(query, s_n) < Th \times EPE$
find $s_x \quad x = 1,..m$ when
$abs(difference\ orientation(query, s_x)) < K$
**end**
send remaining $m$ image to Homology verification step (RANSAC) to find best match.

For our work $Th$ is considered 1.2 to cover bigger region of interest to compensate any unlikely error may happened due to 98% accuracy of $EPE$. Also different values for $K$ that is called here 'view loos' is considered. Figure 5 shows relationship between number of candidate images and $K$ for either a sample for urban area of Chicago that is not really dense region. It can be infer that the number of image is almost the same for this tow regions. So if an approximation of orientation is available, $K = 15$ can be chosen for limiting number of images feeding to Homology verification Although there are some datasets for image retrieval but we preferred to make and use our dataset by the method mentioned before. In this way data for sensors and GPS for query is provided. The quality of images in our database is not really high as others but it seems enough for our research. To generate features we used Vlfeat library [16] and DSIFT as a descriptors. Before applying feature extraction step all images are resized to 300×400 pixel. To exceed our algorithm parallel processing is used to have faster feature extraction procedure. By using proposed method, finding the best match is the same for limited regions and city scale problem. In the other hand proposed method can be applied to a database with thousands of image. The only difference for a city scale dataset and limited dataset is the search for closest coordination with available GSW image. This step that described before is really fast and can be used in real time application. Results show in place with clear view of sky, GPS inaccuracy is low so none of images in dataset will be selected. It means our localization system no longer needs any refinement. In this situation GPS data is reliable and with minimum error. Modifying coordination in such a situation is not reasonable because we only have GSW images almost every 12 meters while GPS accuracy is less than 3 meters. Even if we have more images for example 4 image per meter it won't give use better result for outdoor localization. To evaluate effectiveness of our method some sample images with different cameras have been taken in dense area of Chicago where there was train track in top of us. Also orientation and coordination of camera saved simultaneously. Consider picture 6(a). There are 53 candidate images for this query. Feeding those candidate to Homology verification step that is actually RANSAC algorithm gives us the result in figure 6(c). If we use directly coordination of best match we have the position in fig 6(d) that is really close to our position. Another refinement step can be applied by considering fundamental matrix to get more accurate coordination.

## 7. CONCLUSION AND FURTHER WORK

In this article a method for better localization based on image retrieval is proposed. We tried to utilize this fact that nowadays most of mobile devices such as smartphones are equipped by different inertial sensors. For accurate localization instead of searching among a city scale dataset it would be better to limit search space. We are trying to design a system to help pedestrian especially people with special cases like visual impaired navigate conveniently even in dens area of cities. Knowing that data from GPS can be used along with sensors such as manometer, Gyro and accelerometer for limiting search space made us able to modified Coordination in even difficult condition like sample image is shown above. Although it has reasonable result but it fails in some of samples especially when the quality of images are not good in dataset. In future we are going to use recent feature extraction method like ASIFT to evaluate performance. Also, we will use regular methods like TF-IDf and inverted file approach and try to use GPS data to evaluate results. It is clear that our method performance is high in terms of accuracy because the number of candidate images are limited and knowing that most of image retrieval approaches utilize the same homology verification step as a last step.

Modifying coordination by fundamental matrix between query and best match is another steps that can be taken in account. The ultimate goal of our research is finding an approach to help more accurate device for localization. This research is primary step toward our goal and may need other techniques such as visual odomtry as a complementary method.

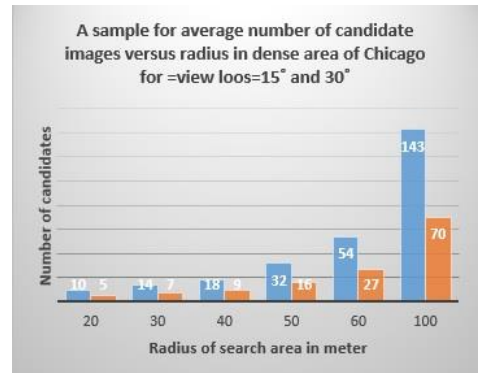

Figure 5.a

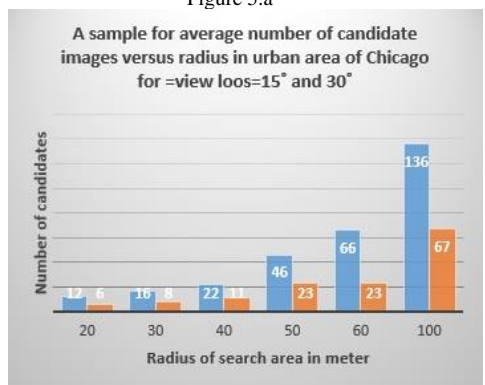

Figure 5.b

## 8. REFERENCES

[1] G.Schroth, R. Huitl, D. Chen, M. Abu-Alqumsan, A. Al-Nuaimi,and E. Steinbach. Mobile visual location recognition. IEEE Signal Processing Magazine, 28(4):77–89, 2011.
[2] A. R. Zamir, A. Darino, and M. Shah, "Street view challenge: Identification of commercial entities in street view imagery," in 2011 10th International Conference on Machine Learning and Applications (ICMLA), vol. 2, pp. 380–383, IEEE, 2011.
[3] D. Lowe. Distinctive image features from scale-invariant keypoints, International Journal of Computer Vision, 60(2):91–110, 2004.
[4] H. Bay, A. Ess, T. Tuytelaars, L. Van Gool, "SURF: Speeded Up Robust Features", Computer Vision and Image Understanding (CVIU), Vol. 110, No. 3, pp. 346-359, 2008.
[5] J. Zhang, A. Hallquist, E. Liang, and A. Zakhor, location-based image retrieval for urban environments, 18th IEEE conference in image processing, 2011.
[6] G. Schindler, M. Brown, and R. Szeliski. City-scale location recognition. In Proc. IEEE Conf. on Computer Vision and Pattern Recognition (CVPR), volume 0, pages 1–7, 2007
[7] G. Reitmayr and T. W. Drummond. Going out: Robust tracking for outdoor augmented reality. In Proc. Intl. Symp. on Mixed and Augmented Reality (ISMAR), pages 109–118, 2006.
[8] G. Reitmayr and T. Drummond. Initialisation for visual tracking in urban environments. In Proc. Intl. Symp. on Mixed and Augmented Reality (ISMAR), pages 161–160, 2007.

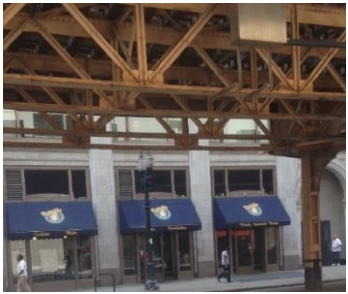 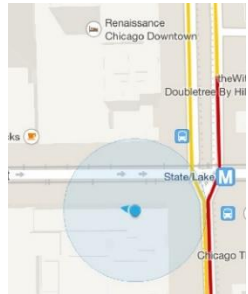 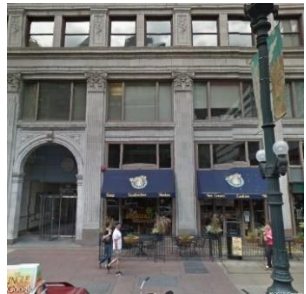 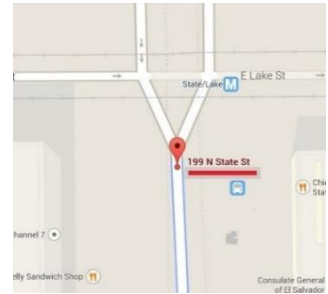

Figure 6.  (a) query　　(b) Coordination of query　　(c) Best matched　　(d) Modified position

[9] Donoser M, Bischof H (2006) Efficient maximally stable extremal region (MSER) tracking. In: Proceedings of conference on computer vision and pattern recognition, pp 553–560.
[10] G. Schroth, R. Huitl, D. Chen, Mobil visual recognition, IEEE SIGNAL PROCESSING MAGAZINE, p77-89, JULY 2011.
[11] J. Sivic and A. Zisserman, "Video Google: A text retrieval approach to object matching in videos," in n Proc. IEEE Int. Conf. Compter Vision, Nice, France, Oct. 2003, pp. 1470–1477.
[12] I. Witten, A. Moffat, and T. Bell, Managing Gigabytes: Compressing and Indexing Documents and Images. San Mateo, CA: Morgan Kaufmann, 1999.
[13] J. Philbin, O. Chum, M. Isard3, J.Sivic, A.Zisserman Improving Particular Object Retrieval in Large Scale Image Databases, IEEE Conference on Computer Vision and Pattern Recognition, CVPR 2008.
[14] M. A. Fischler and R. C. Bolles. Random sample consensus: A paradigm for model fitting with applications to image analysis andautomated cartography. Communications of the ACM, 24(6):381–395,1981.
[15] Tao Guan, Yunfeng He, Juan Gao, Jianzhong Yang, and Junqing Yu ,On-Device Mobile Visual Location Recognition by Integrating Vision and Inertial Sensors, IEEE transactions on multimedia,vol. 15, No. 7, Novemner 2013.
[16] http://www.vlfeat.org.
[17] M. Sharif a, A. Stein a, Ernst M. Schetselaar, Integrated approach to predict confidence of GPS measurement.
[18] M.Salarian, H.Hassanpour. "A New Fast No Search Fractal Image Compression in DCT Domain.", Conf Proc international conference on machine vision, pp.62-66, 2007.